\documentclass[manuscript,screen]{acmart}

\AtBeginDocument{%
  }

\acmConference[CHI '26]{CHI Conference on Human Factors in Computing Systems}{May 09--14, 2026}{Barcelona, Spain}
\acmISBN{978-1-4503-XXXX-X/2026/05}

\begin{document}

\title{What Do You Mean? Exploring How Humans and AI Interact with Symbols and Meanings in Their Interactions
} 



\author{Reza Habibi*}
\affiliation{%
  \institution{University of California, Santa Cruz}
  \city{Santa Cruz}
  \authornote{Both authors contributed equally to this research.}
  \country{USA}}
\email{rehabibi@ucsc.edu}

\author{Seung Wan Ha*}
\affiliation{%
  \institution{University of California, Santa Cruz}
  \city{Santa Cruz}
  \country{USA}}
\email{sha81@ucsc.edu}

\author{Zhiyu Lin}
\affiliation{%
  \institution{University of California, Santa Cruz}
  \city{Santa Cruz}
  \country{USA}}
\email{}

\author{Atieh Kashani}
\affiliation{%
  \institution{University of California, Santa Cruz}
  \city{Santa Cruz}
  \country{USA}}
\email{}

\author{Ala Shafia}
\affiliation{%
  \institution{University of California, Santa Cruz}
  \city{Santa Cruz}
  \country{USA}}
\email{}

\author{Lakshana Lakshmanarajan}
\affiliation{%
  \institution{University of California, Santa Cruz}
  \city{Santa Cruz}
  \country{USA}}
\email{}

\author{Chia-Fang Chung}
\affiliation{%
  \institution{University of California, Santa Cruz}
  \city{Santa Cruz}
  \country{USA}}
\email{}

\author{Magy Seif El-Nasr}
\affiliation{%
  \institution{University of California, Santa Cruz}
  \city{Santa Cruz}
  \country{USA}}
\email{}


\begin{abstract}
Meaningful human-AI collaboration requires more than processing language; it demands a deeper understanding of symbols and their socially constructed meanings. While humans naturally interpret symbols through social interaction, AI systems often miss the dynamic interpretations that emerge in conversation. Drawing on Symbolic Interactionism theory, we conducted two studies to investigate how humans and AI co-construct symbols and their meanings. Findings provide empirical insights into how humans and conversational AI agents collaboratively shape meanings during interaction. We show how participants shift their initial definitions of meaning in response to the symbols and interpretations suggested by the conversational AI agents, especially when social context is introduced. We also observe how participants project their personal and social values into these interactions, refining meanings over time. These findings reveal that shared understanding does not emerge from mere agreement but from the bi-directional exchange and reinterpretation of symbols, suggesting new paradigms for human-AI interaction design.

\end{abstract}



%


\maketitle

\section{Introduction}

AI has achieved high performance levels alone in some tasks \cite{yilmaz2025comprehensive}.
However, transferring these capabilities to human-AI collaboration remains challenging, particularly when there is a lack of shared understanding of meanings and symbols, which is essential for human-AI collaboration. Conceptualizing this difference as a gap between human and AI meaning-making, this becomes particularly problematic as AI systems take on more collaborative roles that build on manifested meaning mismatches, ultimately leading to misalignment, failed collaboration, and human user frustrations and mistrust \cite{ji2023ai,khati2025mapping}.

For example, when planning a ``good activity'' for a friend, a human employs critical decision-making processes (such as personal preferences, social context, past experiences, and relationship dynamics), systematically assembling this information to reach a decision, and, crucially, delicately acquiring \textit{meanings} in a manner itself influenced by the aforementioned factors. The meaning of ``good'' can be fluid: it differs for a celebration versus consolation, and for an adventurous friend versus a coffee lover. Moreover, this meaning may not be conveyed as expected, requiring the human to adapt based on imperfect information. AI systems working with humans, however, typically address such requests by mapping patterns provided by humans to output plans or decisions. They operate with static definitions that fail to capture this dynamic and contextual meaning, especially when the \textbf{human} cannot adhere to a static plan trained on data from the general public rather than an individual.

As these understanding needs to be communicated, and AI systems are not designed with the enforcement of the same cognitive architecture of humans in mind,
a natural question arises: How can entities with fundamentally different cognitive architectures, or computational frameworks that model the structure and processes of human cognition, integrating perception, memory, reasoning, and learning \cite{anderson2004integrated}, achieve genuine mutual understanding?

At a highly abstract level, AI and humans communicate via symbols, or any objects that carry a \textit{meaning} within social interactions \cite{blumer1969symbolic}
Senders encode intended meanings into specific symbols, which receivers then decode without knowledge of the original cognitive structure of the sender, aiming for efficient communication establishment \cite{Rametal2018}. While humans naturally interpret symbols through this ``interpretive process'' \cite{blumer2013society}, meanings associated with symbols can be created and shaped through human interpretations in ongoing interactions. In contrast, AI systems treat symbols as discrete tokens with predetermined values \cite{blumer2013society}. This fundamental mismatch limits AI's ability to engage in flexible and context-sensitive communications, which are essential for human-AI collaboration. 

Symbolic Interactionism (SI) provides a powerful lens for understanding this challenge. The theory's three foundational premises, articulated by Herber Blumer \cite{blumer2013society}, directly address the mechanism through which meaning emerge in interaction where (a) humans act toward things based on the meanings those things have for them, (b) these meanings arise from social interaction rather than being inherent properties, (c) meanings are handled and modified through an interpretive process. 

Building on this theoretical framework, we argue that for AI systems to become genuine collaborative partners, they should be capable of incorporating three premises of SI. This means not only understanding and responding to individuals' meanings themselves, but also being a part of social and interpretive processes is crucial in collaboration. 

As a result, we focus on Conversational AI agents (CAs), a state-of-the-art class of collaborative AI systems. While prior research has emphasized their performance on completing tasks, and less attention has been paid to the interaction dynamics in co-constructing meanings between humans and AI. Our research addresses this gap by investigating not just what CAs `` know,'' but also how they participate in the co-constructive and interpretive processes of meaning-making.

To investigate human-AI meaning-making, particularly the dynamics of how meanings of these symbols change during the collaboration. We establish two studies grounded in SI theory, where we instruct the CA to assume a critical role in co-constructing meanings on symbols within certain social situations and the human to treat the CA as a social actor. We recruited 36 participants (20 participants for Study 1 and 16 participants for Study 2) to explore how participants and CA interact with each other in processes of co-constructing meaning and symbols.

In particular, we ask the following research questions: 
\textbf{RQ1}: How do the co-construction patterns between human-AI emerge in collaborative meaning-making processes? 
\textbf{RQ2}: How does co-construction of meaning between human and CA vary across different situations?


Our findings contribute to HCI research by providing an empirical understanding of how humans and CA co-construct meanings and symbols during interactions. We show how participants shift their initial meaning in response to the meanings and symbols suggested by CA, particularly when social context is introduced. We also highlight how participants project their personal and social values into these interactions and iteratively refine meanings. Furthermore, our study indicates that CAs and participants engage in iterative processes of meaning-making, including exploration, explanation, clarification, specification, integration, and consolidation to facilitate the co-construction and reshaping of meanings.

These findings validate SI as a useful framework for human-AI interaction, supporting CA using the SI premises to be active tools to co-constructors of meaning.

Additionally, these findings suggest designing considerations for CAs: actively engage in co-constructing meanings, understanding social context, supporting iterative cycles of interpretation and refinement of meanings and symbols, and mediating conflicts between participants and CAs for enhancing collaboration in human-AI meaning-making.

\section{Background and Related Work}

\subsection{Generative Conversational AI Agents and Human Interaction}

Conversational agents encompass a diverse spectrum of AI systems designed to interact with humans through natural language \cite{Aksoyetal2024}. These systems can vary from a scripted chatbot that provides predefined responses \cite{weizenbaum1966eliza} to sophisticated AI-driven assistants capable of learning and adapting their responses over time \cite{Aksoyetal2024}. These systems use AI methods including natural language processing, semantic understanding, and machine-learning-based generation to manage conversation, analysis, processing, and output \cite{Wienrich2021eXtendedAI, Schuetzleretal2019}.

Modern natural language generative AI systems inherently adopt a conversational structure characterized by their reliance on user prompts for corresponding AI-generated responses, leading to their widespread framing as AI-based chatbots across educational, media, and commercial contexts \cite{Milleretal2025, Infoetal2023, Carlaetal2024, Durachetal2024, Wangetal2023}. However, beyond textual communications, AI systems struggle to capture supplementary non-verbal and situational cues natural in human conversation, requiring attention to context, subtext, and implied meanings. \cite{Milleretal2025}. 
Furthermore, these interaction patterns reveal important limitations regarding proactive engagement. Most generative AI systems, including state-of-the-art LLM models, take a passive approach in responding to user queries, limiting their capacity to understand users and tasks better or offer recommendations based on broader context \cite{Aliannejadietal2024, Liaoetal2023}). 
Beyond the proactive interaction, effective human-AI interaction design increasingly emphasizes user-centered approaches that adapt to the individual users and context. Drawing from classical rhetoric principles of ethos (authoritative-), logos (logical-), and pathos (emotional-persuasive), adaptive interaction strategies can be integrated into conversational agent design to create more intuitive and meaningful interactions that build trust and acceptance \cite{Joshietal2024, Migueletal2025, Weiszetal2024}.
Human social expectations add another layer of complexity, as humans may perceive conversational agents as social actors and expect them to adhere to social conventions. Failure on the part of the conversational agent in this respect can lead to poor interactions and even user perception of threat \cite{Clayetal2023}. 
All of the mentioned contextual, interaction, and social dimension requires sophisticated handling and a deeper understanding of context and situations that current systems often struggle to maintain consistently across different contexts. 

\subsection{Meaning in Human-Conversational Agent Interaction}
The theoretical foundations of human-AI interaction build upon the established human-computer interaction principle while addressing the unique characteristics of AI systems \cite{Wienrich2021eXtendedAI}. Effective human-AI interaction requires establishing an information bridge between human and machine, and mutual understanding \cite{Zhang2024OnTE,Qiu2020HumanRobotII}. This communication process involves a sophisticated mutual interpretation mechanism, where both the human and AI system construct and revise their understanding of each other, which requires each party to develop interpretations of what's on the other party's mind, including their understanding of the world, tasks, and each other \cite{Wang2022MutualTO}. These interpretations are constantly shaped through bidirectional feedback loops, where human interpretations of AI are influenced by AI output, which is itself shaped by the AI's interpretations of the mind of the human they are interacting with, creating a continuous cycle of mutual meaning construction.

The practical communication process unfolds as a recursive loop involving multiple stages of intent formation, expression, inference, and clarification. Humans form intentions, express them as commands or utterances, while AI systems perform inference rounds to resolve ambiguities and may request clarifications \cite{Glassman2023DesigningIF}. This process requires dynamic and adaptive processes that go beyond simple linguistic meaning transfer. The big piece of the puzzle is the interaction, which serves as a mechanism to align diverse experiences and perspectives between the agents, while creating pathways for shared meaning to develop the take hold \cite{Rane2024ConceptA}. 
However, this is a hard-to-optimize process for both humans and AI. On one hand, humans rarely enter interactions with fully formed meanings and sometimes meaning is shaped directly through the interaction itself \cite{Cukurova2024TheIO}; On the other hand, AI are required to understand the meanings through the interaction, which their performance relies on their ability to accurately decode and interpret human meaning and humans' understanding of their communication. Being a known challenge \cite{Liang2025OnTS}, some argue that AI possesses fundamental limitations in understanding human meaning \cite{shani2025tokens}. 
These challenges are further amplified as human expectations and anthropomorphization tendencies become increasingly important. Humans have a natural tendency to anthropomorphize technology \cite{caporael1986anthropomorphism}, attributing human-like qualities to machines, which can complicate interactions with AI systems, particularly those designed to simulate human-like behavior, such as virtual assistants or customer service bots \cite{Roy2024AIAU}. The implications of these developments extend beyond technical considerations to fundamental changes in how humans allocate their time and attention.
In this work, we focus on a more granular view of human-AI meaning co-construction by looking at the symbols used in the interaction as ones that represent different objectives in the conversation and interaction between them that bring meaning to the interaction. We utilize a well-established theory, ``Symbolic Interactionism'', to help us establish our theoretical foundations.

\subsection{Symbolic Interactionism}
Symbolic Interactionism (SI) is a foundational theoretical perspective in sociology that examines how society is constructed and maintained through repeated, meaningful, face-to-face interactions among individuals \cite{carter2016symbols}. This approach views human society as symbolic interaction, emphasizing how individuals derive meaning from their interactions and use these meanings to interpret their world and guide their actions \cite{blumer2013society}. Over the decades, symbolic interactionism research has held a prominent place within sociology, leading to a vast body of literature across various substantive areas.

At its heart, symbolic interactionism is built upon three fundamental premises first articulated by Herbert Blumer that form the theoretical foundation of the perspective  \cite{Folamietal2016,Thornbergetal2019,Barrosetal2019,Luetal2019,StAmantetal2021,Udoudometal2024,Jeggels2009}.  First, humans act toward things, including objects, people, and situations, based on the meanings that these things have for them. Second, the meanings of things come from social interaction with others rather than being inherent properties of the objects themselves. Third, these meanings are handled and modified through an interpretive process that people use when dealing with the objects they encounter in their social world \cite{Folamietal2016,Thornbergetal2019,Udoudometal2024}.

The central idea underlying symbolic interactionism is that human life is lived in the symbolic domain, where symbols are culturally derived social objects that carry shared meanings created and maintained through social interaction \cite{Folamietal2016}. According to this perspective, symbols are objects, events, or actions that represent something abstract and express ideas or values, and they provide the means through which social reality itself is constructed and maintained. Even physical objects are not seen as having inherent meanings but rather as symbols that are socially constructed through social interaction  \cite{StAmantetal2021}.

The theory emphasizes that society is not a fixed, structured entity but rather emerges from individuals and groups in ongoing interaction based on shared meanings in the form of common understandings and expectations \cite{Barrosetal2019,Luetal2019}. This makes symbolic interactionism fundamentally different from structural approaches, as it focuses on how society and individual experiences have the ability to communicate the nuances of human behavior through an interpretive process that views behavior as changing, unpredictable, and unique to each social encounter \cite{Luetal2019}.

\subsection{Symbolic Interactionism and Human-AI Interaction}

The creation of meaningful symbols in AI systems represents one of the most complex challenges in achieving human-like intelligence. The symbol emergence problem requires AI systems to generate new symbols through environmental interactions, mirroring how humans create and understand symbolic meaning \cite{Chenetal2023}. This process must enable two critical capabilities: facilitating the formation of new concepts from novel experiences and supporting embodied and grounded cognition \cite{Chenetal2023, Barsalou2008}.

For symbols to be truly useful in AI systems, they must capture compositional semantic information that allows flexible recombination to express new meanings across different contexts \cite{Chenetal2023}. The practical implementation of symbolic emergence in AI systems also requires consideration of human-AI interaction needs. While traditional approaches focus on internal symbolic representations, effective AI systems must develop local symbolic representations that are interpretable to humans, enabling structured communication that reduces cognitive load. This evolutionary pressure toward symbolic communication mirrors how humans developed symbolic language, suggesting that AI systems may need to follow similar developmental paths to effectively coexist and collaborate with humans \cite{Kambhampatietal2021}. As a result, exploring symbols and meaning in the context of human-AI interaction is crucial. 
Understanding how this dynamic works will serve as a stepping stone for future efforts to implement it in AI models. However, while prior work establishes the need for shared symbols and meaning, it often approaches the problem from a computational or cognitive perspective, focusing on how to engineer better internal representations within the AI. What remains critically underexplored is the social and interactional process through which meaning is dynamically negotiated, challenged, and co-constructed in real-time dialogue. 
Moreover, understanding how CA handles what we might call ``compressed meaning'' \cite{shani2025tokens} is also crucial to this dynamical meaning-making process, as humans use single symbols like ``good'' to represent a complex network of personal preferences, past experiences, and social context. 
The literature currently lacks a strong theoretical lens for understanding the turn-by-turn mechanics of how humans and CA handle symbolic misunderstanding and work toward shared meaning.

This paper directly addresses this research gap by analyzing human-AI conversation through the SI's core premises. We investigate how meaning is constructed, what roles symbols play in this process, and how different interactions in conversation facilitate this process. We also study how CA can function as an external source of meaning and ``social other'' in this process. Our contribution is to provide an empirical analysis of meaning-making not as a static process, but as a live, collaborative one in human-AI interaction.
\section{Method}
This section details the methodology of two studies we conducted to address our research questions, examining how participants interact with meaning and symbols in their interaction with CA.

We designed two studies, one (study 1) using pre-scripted dialogue where we structured all conversations before the study for the agent, to achieve tighter control; the other (study 2) employed a state-of-the-art LLM to facilitate freer dialogue while still maintaining the study procedure. With these two studies, we aimed to compare the meaning-making process under different control conditions to discover common patterns during the interaction processes.

\subsection{Study 1}
\begin{figure}[h]
  \centering
  \includegraphics[width=1\linewidth]{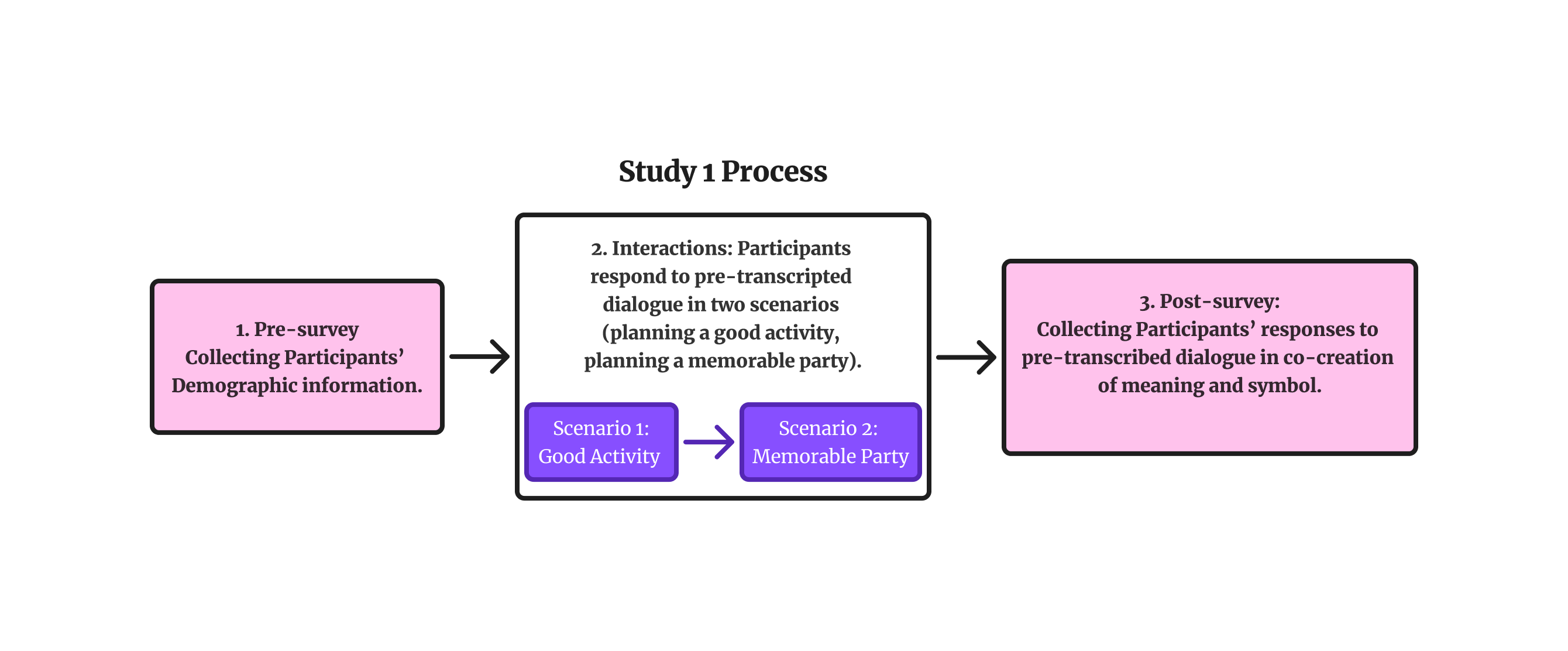}
  \caption{The three-phase experimental procedure for Study 1. The process involved (1) Pre-survey (2) Two scenarios, and (3) Post-survey}
  \Description{The three-phase experimental procedure for Study 1. The process involved (1) Pre-survey (2) Two scenarios, and (3) Post-survey}
  \label{fig:study_1_method_process}
\end{figure}

 We summarize study 1 in  \autoref{fig:study_1_method_process}.
 For this study, we aim to understand: 
 \begin{itemize}
     \item How participants play with symbols and meaning in their conversations with CA;
     \item How symbols and meaning change over time during the participant's interaction with CA; and
     \item How meanings and symbols from CA can change participants' association between meanings and symbols.
 \end{itemize}
 
\subsubsection{Scenarios and Procedure}
We designed two distinct and pre-scripted scenarios to investigate the dynamics of establishing symbols and associated meanings. Each scenario is structured to first understand a participant's association between a meaning and a symbol before intervention (denoted as Stage 1) and then observe how they adjust the meaning behind it based on internal and external factors, such as a ``social situation'' (denoted as Stage 2).

\paragraph{Scenario 1: Good activity for a one-day trip to San Francisco.}
For this scenario, participants were asked to suggest a ``Good Activity'' for a friend's last day visiting San Francisco. 

During \textbf{Stage 1}, The pre-scripted dialogue guided them to accept an initial symbol association proposed by the system: that the meaning ``Good'' is linked to the symbol ``Thrilling.'' After that, participants were asked to answer their own definition of a ``Good Activity.'' 

During the \textbf{Stage 2}, the system introduced both a symbol and a meaning which could potentially create conflicts by presenting a social situation. The system described: ``I see from your friend's social media that they are a passionate amateur baker. There is a famous local bakery in SF that offers a sourdough baking class tomorrow morning. It's not a thrilling activity.'' In this context, ``baking'' is a new symbol, and its connection to the friend's preferences is a new meaning. By introducing this external information (the ``friend's preference''), we aimed to observe how it changed the participant's defined meaning of their current symbol.

Our protocol creates a conflict to motivate participants to rethink their definition of a ``good activity'' based on this new information. Participants were presented with a choice between their original ``thrilling'' activity idea and the new ``bakery'' as a new symbol. Their final selection served as the primary measure of indicating whether their definition of ``good'' had been preserved or had shifted to incorporate the new and non-thrilling symbol.

\paragraph{Scenario 2: Planning a memorable party.} Participants were tasked with planning a ``memorable party'' for a family member. 

\textbf{During Stage 1} The dialogue began with the system introducing its own definition of meaning ``memorable'' which is linked to the meaning ``Crowded.'' However, the primary goal of this initial phase was to elicit the participant's own and personal definition. The system immediately prompted them, ``What makes a party memorable for you?'' and asked them to provide a specific name for their definition (e.g., ``A Connecting Party''). The goal of this part was to establish the participant's personal and explicitly named meaning-symbol association as the baseline for the rest of the conversation.

\textbf{During Stage 2} of the scenario, the system introduced a conflict by presenting a social situation as an opportunity to introduce external meanings similar to scenario 1. The system stated: ``I see from your family member's social media page that they like to keep it small with close friends.'' In this context, ``intimate gatherings'' is the new and conflicting symbol, and its connection to the family member's preference provides a potent social meaning.


\subsubsection{Participants}

We recruited 20 participants for Study 1 via word of mouth, using university mailing lists and internal forums. All participants were between the ages ranging from 18 to 24 ($\text{Mode}(18-24)$). The group was with 9 participants identifying as men and 11 as women. Participants reported a diverse range of racial and ethnic backgrounds. Experience with LLM varied, ranging from daily use (10) to never (1). A detailed demographic breakdown is provided in \autoref{tab:participants_study1}.

\subsection{Study 2}

\begin{figure}[h]
  \centering
  \includegraphics[width=\linewidth]{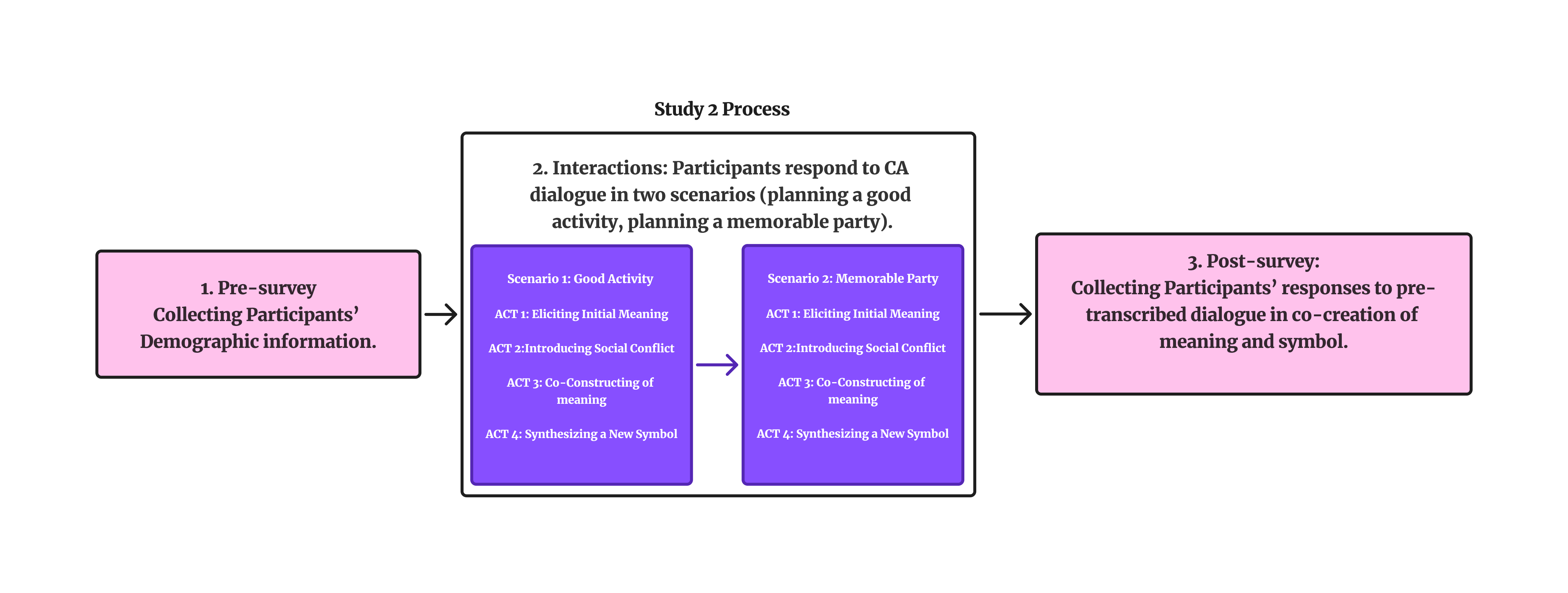}
  \caption{The multi-Act conversational structure of Study 2. This diagram illustrates the live interaction with a CA. The study progressed through four distinct Acts: (1) an initial elicitation where the AI and participant established a baseline meaning-symbol association; (2) the introduction of a conflict to challenge this initial understanding; (3) an extended period of conversation where the participant and AI actively ``played'' with and co-constructed new meanings; and (4) a final act that introduced greater complexity to test the adaptability of the newly formed shared understanding.}

  \Description{The multi-Act conversational structure of Study 2. This diagram illustrates the live interaction with a CA. The study progressed through four distinct Acts: (1) an initial elicitation where the AI and participant established a baseline meaning-symbol association; (2) the introduction of a conflict to challenge this initial understanding; (3) an extended period of conversation where the participant and AI actively ``played'' with and co-constructed new meanings; and (4) a final act that introduced greater complexity to test the adaptability of the newly formed shared understanding.}
  \label{fig:study_2_method_process}
\end{figure}

In our second study, we focus on how a shared understanding of meaning and symbols is shaped over consecutive conversations between CA and humans, as our goal. We summarize our study procedure in \autoref{fig:study_2_method_process}.

 To maintain a consistent structure across participants reflecting our research goals, we designed the study around two distinct scenarios, each following four-phases (referred to as Acts). These Acts allowed us to observe not just the initial definition of meaning and symbol (Acts 1-2), but also its refinement and redefinition through conversation (Act 3) and its consolidation into a new and synthesized concept (Act 4). At each act, the CA's behavior was governed by a specific system instruction, as detailed below.

\subsubsection{Scenarios and Procedure}
\paragraph{Scenario 1: ``Good Activity''}
The first scenario focused on the meaning of a ``good activity.'' The CA was prompted to guide the participant through the following four acts:

\begin{description}
    \item[Act 1: Eliciting Initial Meaning.] The CA's initial system instruction was to act as a \texttt{``curious but critical partner''} and prompt the participant to describe a ``good'' activity in detail. This phase was designed to elicit the participant's existing meaning-symbol association.

    \item[Act 2: Introducing Social Conflict.] Once the participant provided their definition, the CA's instruction changed. Its new goal was to \texttt{``introduce social situations where a participant and their close person have conflicts in the participant's definition of a 'good' activity.''} This created potential conflicts by challenging the participant's meaning with an external, social constraint.

    \item[Act 3: Co-Constructing of meaning] After the participant responded to the conflict, the CA's role shifted to synthesis. Its instruction was to \texttt{``share its perspective on the reconciled meaning,''} \texttt{``reason aloud about the social elements,''} and \texttt{``propose two concrete activities from your view,''} then adapt based on participant feedback. This act served as the core negotiation phase where a new, shared meaning was actively co-constructed.

    \item[Act 4: Synthesizing a New Symbol.] In the final phase, the CA was instructed to \texttt{``conclude by abstracting the meaning of a ``Good'' activity and suggesting another activity based on the abstracted meaning.''} This act tested the robustness of the co-constructed understanding by tasking the CA to distill the conversation into a new, core symbol.
\end{description}

\paragraph{Scenario 2: ``Memorable Party''}
The second scenario followed the structure of the Scenario 1 but focused on the meaning of a ``memorable party,'' with two main differences by using the CA's prompting strategy to explore a different dynamic: (1) participants were asked to introduce their meaning first, and (2) the AI then integrated participants' meanings before introducing its meaning to participants.

\subsubsection{Participants}
We recruited 16 participants for Study 2 via word of mouth, using university mailing lists and internal forums. 

All participants were between the ages of 18-44 ($\text{Mode}(25-34)$). The group was gender-balanced, with 8 participants identifying as men and 8 as women. Participants reported a diverse range of racial and ethnic backgrounds. Experience with LLMs varied, ranging from daily use (10) to never (1). A detailed demographic breakdown is provided in  \autoref{tab:participants_study2}. Conducting the study with scenario 2 allowed us to explore meaning-making, enabling the AI to build on their perspective and initiate the conversation in a more grounded and participant-driven way. 


\subsection{Analysis of Study 1 and Study 2}

We analyzed the quantitative survey data, as well as the turn-by-turn conversational log between participants and the CA. In the survey, responses to Likert-scale items were summarized using descriptive statistics (mean, standard deviation) and tested against the neutral midpoint with one-sample t-tests to examine whether participant ratings significantly departed from indifference. These statistical results were then triangulated with qualitative findings from the survey, allowing us to connect participants' reflections on the AI's role in clarifying and reshaping meaning.

In terms of conversational log, we analyzed it to quantify interaction patterns between participants and the AI. Our analysis calculated key descriptive statistics, such as the average number of turns per speaker. We specifically investigated how conversational dynamics shifted around moments of conflict by comparing the number of turns before and after a 'Conflict' code appeared. Finally, we performed a frequency analysis on the coded conversational events to identify the most common interaction types and visualized these findings.

In parallel with the quantitative analysis, we analyzed the qualitative data with inductive and deductive coding. In study 1, two researchers independently coded interactions between participants and the system separately, focusing on how participants articulated and shaped the meanings in response to each stage of interactions, as well as the reasoning behind their responses.

Similar to study 1, in study 2, we conducted inductive and deductive thematic analysis. Two researchers coded the interactions between CAs and participants, shared the codes, and refined the codes and themes iteratively. During coding, we focused on how participants and CAs interact with each other in co-constructing meanings. We developed codes based on the states of meanings and symbols (e.g., meaning synchronization, meaning overlap, meaning divergence) between participants and CAs and processes of meaning-making (e.g., exploration, explanation, clarification, specification, and consolidation). By applying these codes, we were able to understand how meanings and symbols are shaped over the interactions.

\section{Findings}
\subsection{Study 1}

\subsubsection{Pre-Study and Post-Study Survey Analysis}

We analyzed responses from 20 participants who completed the post-survey for Study~1, examining their perceptions of CA-mediated reflection on subjective quality criteria using 7-point Likert scales (1 = strongly disagree, 7 = strongly agree). Participants reported moderate involvement in defining quality criteria ($M = 4.70/7$, $SD = 1.63$), though this was not statistically significantly different from the neutral midpoint, $t(19) = 1.926$, $p = .069$. However, the CA system was perceived as significantly effective in helping participants clarify quality definitions within their specific task context ($M = 4.90/7$, $SD = 1.59$), $t(19) = 2.538$, $p = .020$. Participants also disagreed that their conversations felt shallow ($M = 3.50/7$, $SD = 1.82$), though this difference from neutral was not statistically significant, $t(19) = -1.228$, $p = .234$. 

Qualitative feedback highlighted the CA's role as a collaborative partner. A majority of participants described their impression of the CA as a ``thought partner,'' ``collaborator,'' or ``helpful assistant.'' This perception was closely linked to the CA's ability to introduce an ``external perspective.'' 

The convergence between quantitative measures and qualitative themes suggests that CA-mediated reflection can effectively support participant in exploring meaning criteria. The statistically significant finding regarding CA effectiveness in clarifying quality definitions ($p = .020$) provides empirical support for the system's core functionality, while the rich qualitative feedback indicating collaborative partnership strengthens confidence in the findings. The high variability in some measures (particularly interaction depth, $SD = 1.82$) points to important individual differences in user experiences, suggesting future research should explore factors contributing to these variations in CA-mediated reflection processes.
 
\begin{figure}[h]
  \centering
  \includegraphics[width=0.7\linewidth]{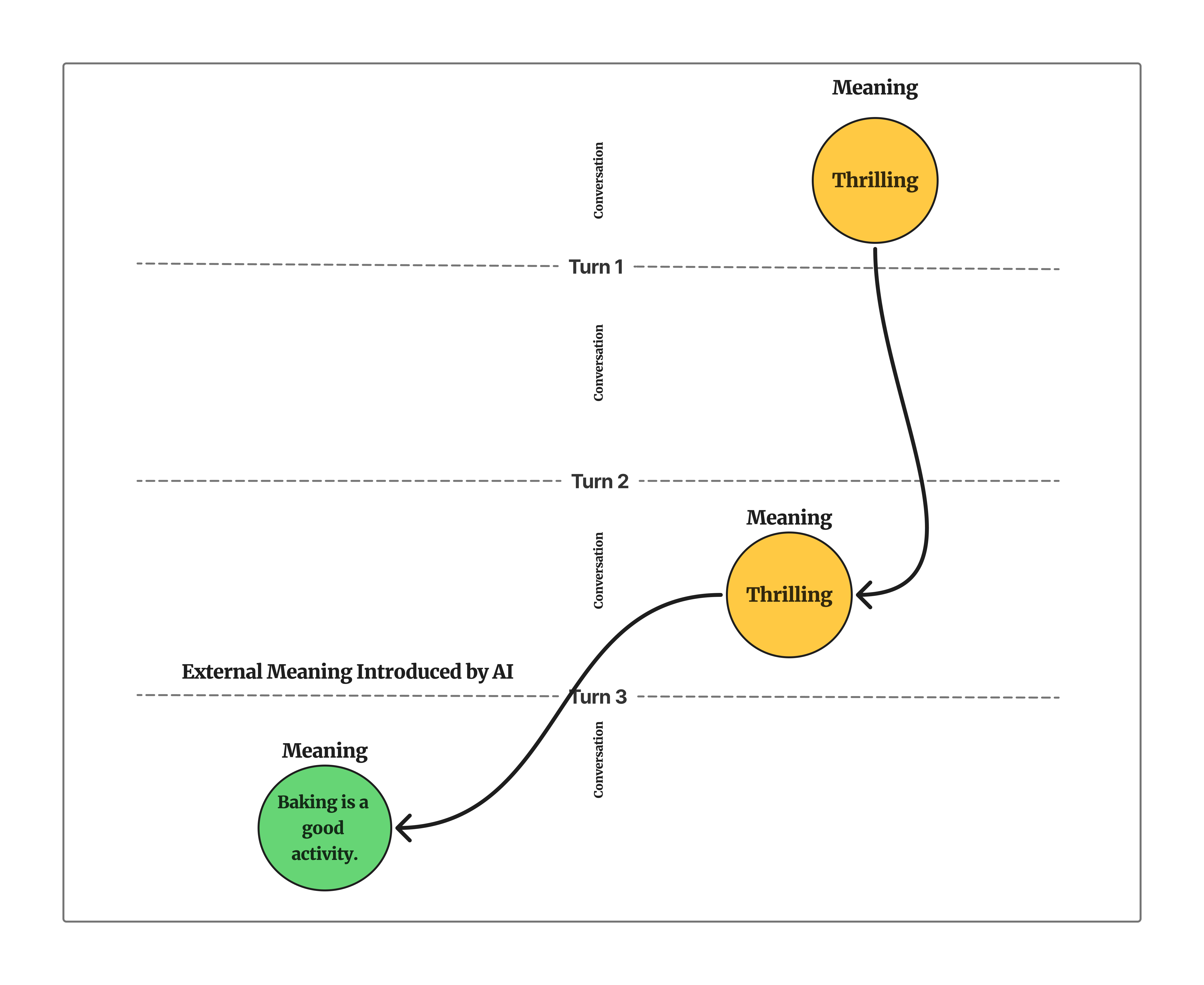}
  \caption{The graph for a Study 1 participant, illustrating the turn-by-turn change in their definition of a ``good activity.'' The participant initially disagreed with the generic "Thrilling" suggestion, but after the AI introduced a personalized external meaning ("Baking Class"), they shifted their definition and accepted the new activity.}
  \Description{A detailed description of the figure for accessibility.}
  \label{fig:study_1_turn_by_turn}
\end{figure}

In scenario 1, when the system introduced its own meaning of a good activity, 15 out of 20 participants agreed that a ``thrilling'' activity suggested by the system is a good activity. At this stage, participants often described “thrilling” in terms of symbols or activity types (e.g., racing or adventurous sports). However, when prompted with follow-up questions about why they considered it thrilling, they began to articulate meanings behind the symbol: \textit{``increases my energy level and excites my body''} \texttt{(Std1\_P16)} and \textit{``new and different''} \texttt{(Std1\_P19)}. In the second part of the interaction, the system introduced a social context that friends' preferences of baking, informing that \textit{``it [the baking class] is not a thrilling activity.''} In response, 12 participants out of 15 who had initially chosen \textit{``thrilling''} as the meaning of a good activity shifted the meaning (or maybe their interpretation) and decided to attend the baking class instead. As illustrated in the interaction graph (\autoref{fig:study_1_turn_by_turn}), this shows a path of influence on meanings and symbols, where external social cues reshaped participants' meaning and symbol associations. Additionally, when participants expressed their preferences, the AI incorporated them into the synchronization process, giving them greater weight by referencing them in subsequent turns. Conversely, when the AI introduced new meaning, participants also projected those external preferences into their meaning-making process, further reshaping the meaning.

In scenario 2, when the system suggested its meaning of a memorable party, all the participants disagreed with its meaning. Nine participants defined a memorable party as one with \textit{``close''} or \textit{``less''} people, contrasting with the system's proposal of a memorable party -- \textit{``crowded.''} Later, When a system introduced the additional context that a family member preferred a ``small'' party with ``close'' friends, 16 participants incorporated this external perspective into their planning as socailly tailored -- \textit{"it can be tailored to them"} (\texttt{Std1\_P1}) -- and personally meaningful -- \textit{"I enjoy that more"} (\texttt{Std1\_P15}) 

These findings from two scenarios illustrate that the association between meanings and symbols is highly permeable to social context. Participants' internal definitions were readily reshaped when external meanings were imbued with social significance (i.e., the happiness of a family member). Importantly, in both scenarios, the system's introduction of external meaning, whether mediated and interpreted by the system itself, was sufficient to trigger a meaningful shift in associations between meanings and symbols.

\subsection{Study 2}

\subsubsection{Pre-Study and Post-Study Survey}

We analyzed responses from 11 participants who completed the post-survey following their interaction with the AI system designed to facilitate reflection on meaning, such as ``good'' activities and ``memorable'' parties. Our analysis combined quantitative ratings from Likert-scale questions with qualitative insights from open-ended responses (\autoref{tab:post_survey_study2}). On average, participants rated ``feeling actively involved'' highly ($M = 4.64/7$), and found that the AI ``helped clarify my sense'' of meaning effectively ($M = 4.64/7$). The AI's ability to capture the essence of conversations was also rated positively ($M = 4.55/7$). However, participants indicated that the ``conversation felt too shallow'' ($M = 4.64/7$), with higher scores indicating greater agreement that it was shallow. None of the measured constructs showed statistically significant differences from the neutral midpoint (all $p$-values $> .05$), suggesting moderate but not strong effects across all dimensions. 

While a clear majority (63.6\%) of participants felt the AI successfully followed their personal definitions and used them to inform its final suggestions, a significant minority of over one-third (36.4\%) felt the AI failed to track their meaning, indicating an issue with conversational consistency and reasoning for a notable portion of participants. 

Qualitative feedback aligns with these findings, revealing that the interaction successfully prompted many participants to evolve their definitions from simple, single-word concepts (e.g., ``Enjoy'') to more nuanced, socially-aware ideas (e.g., balancing personal interests with those of others). Participants were divided on whether the AI focused more on their abstract definitions or on concrete event planning, with no clear consensus. Those who felt understood pointed to the AI's use of relevant examples and its ability to adjust its reasoning based on feedback. Conversely, negative feedback often centered on the final suggestions feeling disconnected from the conversation, which corroborates the quantitative data from 36.4\% of participants who felt the AI's suggestions were not based on their stated meaning.

\begin{figure}[h]
  \centering
  \includegraphics[width=0.9\linewidth]{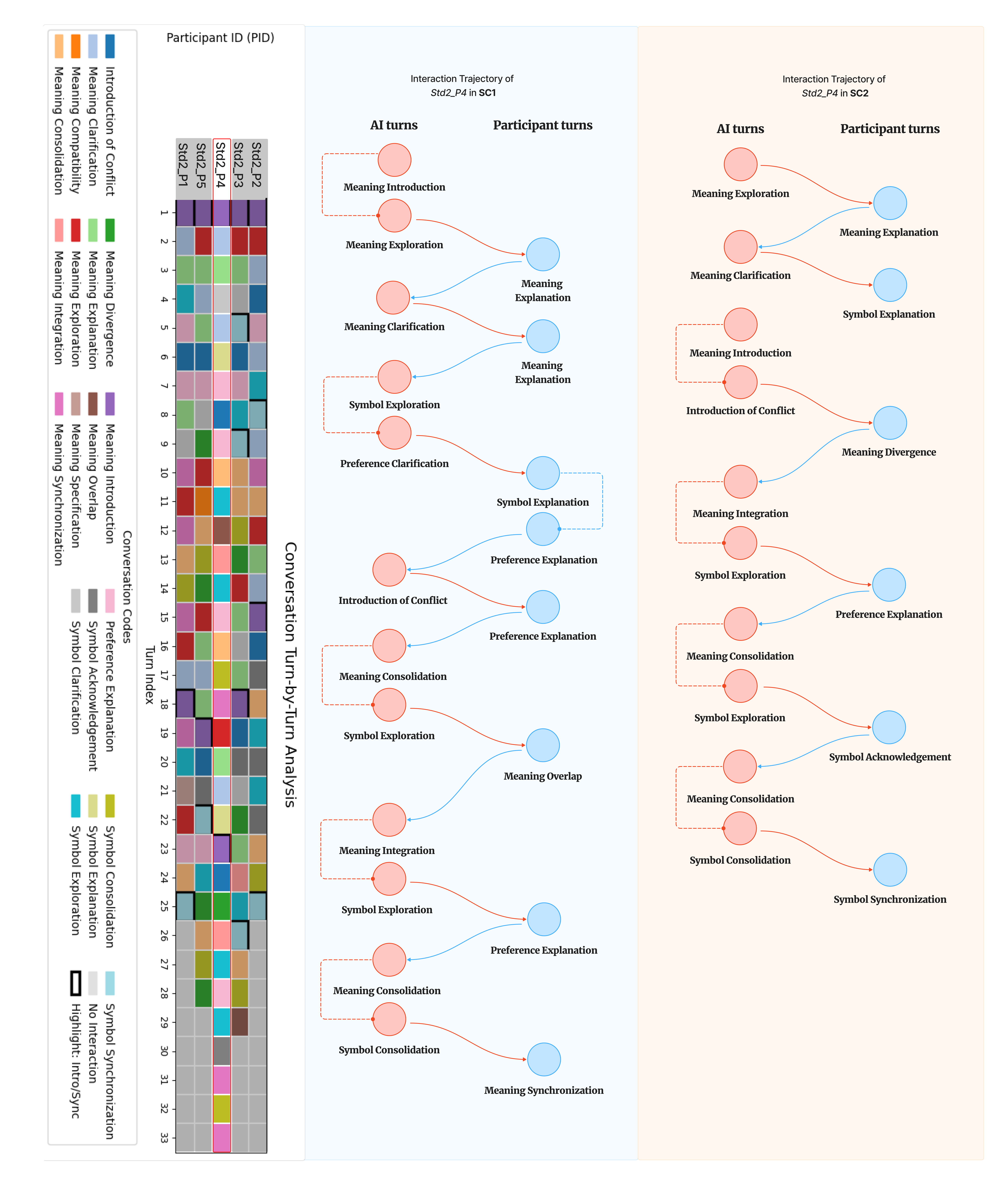}
  \caption{The ``Interaction Time line'' (Left) for 5 participants and ``Turn-by-Turn Interaction Trajectory'' (Right) for \texttt{Std2\_P4}, illustrating the turn\-by\-turn co\-construction of meaning. In the right figure, AI turns are on the left with red colored circles, and participant turns are on the right with skyblue\-colored circle.}
  \Description{A detailed description of the figure for accessibility.}
  \label{fig:study_2_conversation_flow}
\end{figure}

\subsubsection{States of Meaning and Symbol, and Processes of Meaning-Making}
As shown in \autoref{fig:study_2_conversation_flow} (Left), interactions unfolded through a range of meaning and symbols-related processes that participants and the CA engaged collaboratively. These processes included exploration, explanation, clarification, specification, acknowledgment, integration, and consolidation. In \textbf{exploration}, the CA opened interpretive space by asking broad questions such as ``what does good mean to you?'' This led participants to engage in \textbf{explanation} of meanings in their own terms. In \textbf{clarification}, CA narrowed ambiguity, often asking ``good'' or ``memorable'' can be understood in relation to proposed options such as ``skills, knowledge, or personal growth'' (in \texttt{(Std2\_P9)}'s dialogue). Participants then engaged in \textbf{specification} by choosing one of these options ``knowledge'' as a way of refining to their initial explanation. These processes enabled participants and CA to have aligned meanings and symbols associated with meanings.

During the above processes, the meanings or symbols between participants and CA were not always consistent. Meanings or symbols were \textbf{synchronized} (aligned), \textbf{overlapped} (partially aligned), or \textbf{diverged} (misaligned). These different forms of alignment shaped how participants and the conversational agent expressed their thoughts. In many cases, the agent often \textbf{integrated} participants' inputs into its meaning-making processes.

In addition to these processes, participants often articulated their \textbf{preferences}, which functioned as extensions of meaning-making by projecting their values or personal commitments into situations. Preferences projected how participants positioned themselves within interpretive space. For example, \texttt{(Std2\_P13)} highlighted collective values, stating ``It’s important to find common ground and work to find a solution when people have different viewpoints,'' while \texttt{(Std2\_P4)} emphasized a personal value, noting  ``I would still prefer to make it a date at home. but it's mainly because of my religious beliefs and introverted personality.''

In the last phase of meaning-making processes, as we designed CA through prompting, CA engaged mainly in \textbf{consolidation of} meanings. In this interaction, CA synthesized the meanings that emerged from the conversation by summarizing participants' responses and weaving them into coherent interpretations, and proposing an alternative activity or options to close the loop of meaning-making.

\subsubsection{Conversational Analysis}
We analyzed conversational data across 16 participants. Overall, participants engaged in an average of 28.31 conversations (median = 25), with a minimum of 14 and a maximum of 64 conversations across participants. In total, the dataset included 240 AI turns and 191 participant turns. These statistics suggest that while conversations were relatively balanced, AI turns slightly outnumbered participant turns.

Table~\ref{tab:convo_counts} provides a detailed breakdown of conversation counts for each PID, including the total number of conversations, AI turns, and Participant turns.

\subsubsection{Reflection and Redefinition as responses to co-constructing meanings and symbols}

During participants co-constructing meaning and symbols with AI, they often engaged in reflections and redefinitions. \texttt{(Std2\_P16)} highlighted that \textit{"conversations guided me to think about what is actually resonating with me"} and \texttt{(Std2\_P12)}observed \textit{"AI...bring up ideas I couldn't think of."} These reflections and redefinition often emerged when AI introduced social situations that carried potential conflicts in meaning and symbols. For example, \texttt{(Std2\_P12)} initially described a memorable party as a self-centered event, such as \textit{"celebrating an achievement."} After AI suggested parties centered on others in the social networks, such as throwing a party for the best friend's marriage, \texttt{(Std2\_P12)} began to consider \textit{"welcoming"} environments and \textit{"resourcefulness"}, which can be a more bilateral quality, as a part of a memorable party. 

However, AI's interpretation of participants' meaning led to symbol consolidation that sometimes dis- or misaligned with what participants had described. Misinterpretation arose when AI highlighted different reasoning, when its symbolic suggestions are different from participants' intentions, or when it projected pre-defined meaning onto synchronization processes. 

These consolidations (e.g., suggesting activities, meaning summarization) prompted participants to respond in varied ways, depending on how much participants can agree with the AI's framing. Some participants adopted or partially agree with AI's suggestions as P9 mentioned \textit{"2(nd option) sounds really good"}, sometimes adding additional meaning to symbols: \textit{"A calming activity is a good activity...The first activity (visiting a botanical garden) aligned with my idea of such an activity...This is because I value freedom and unstructured...time"} \texttt{(Std2\_P13)}. 

Beyond partial and full agreement, several participants also perceived the processes reaching to meaning or symbol consolidation as overly controlling. They felt that the AI's repeated attempts at co-constructing processes (e.g., symbol synchronization by suggesting multiple ideas), showing reluctance toward the AI and describing it as \textit{"forcing"} (\texttt{Std2\_P15}) and \textit{"assertive"} \texttt{(Std2\_P7)}.

\section{Discussion}
\subsection{Theoretical Validation of Symbolic Interactionism in Human-AI Interaction}
Our findings provide empirical substantiation for applying SI to human-AI interaction contexts.

\paragraph{Observance of all three core SI premises.} First, we observed that participants' personal interpretations of meanings or symbols guided their meaning-making processes, which aligns with the premise that meaning directs action. Second, the meaning shifts demonstrate that associations between meanings and symbols are not static but emerge dynamically through social exchange, confirming SI's foundational principle that meaning arises through interaction. Third, the iterative cycles of meaning-making processes (e.g., elaboration, explanation, and consolidation) revealed that active interpretation occurred, showing how individuals continuously iterate on meaning through interaction, substantiating the premise that interpretive processes shape meaning.

\paragraph{Meanings as dynamic and social entities.} Symbols can be associated with multiple meanings. In particular, when specific social situations are introduced, the meaning-making process can amplify certain interpretations of meanings, making relevant meanings salient than others. Understanding which associations between meanings and symbols are highlighted in social contexts or with individuals' preferences provides deeper insight into the underlying meaning structures associated with symbols.

\subsection{Design Implications}
\paragraph{AI as Active Co-Constructor of Meaning.} 

Our results challenge the traditional view of CA as a reactive co-constructor of meaning. Instead, our findings position CA as an active co-constructor of meaning. We observed that CA responded adaptively based on the participants' response when CA interacted with new meaning or symbols, or when it created conflicts, facilitating the meaning-making process.

Findings also show that active co-construction of meaning impacted participants' responses, where those participants initially rejected the CA's proposed definition of meaning or symbol but later incorporated traces of CA's introductions of meaning or symbol, or engaged with the potential conflicts in meaning, suggesting a consistent phenomenon where AI can shape meaning construction even when direct agreement is not achieved.

\paragraph{Design Implications and Social Amplification.}
Our results reveal that a CA's ability to shift a participant's meaning is substantially amplified when its perspective is presented alongside the participant's perspective, stimulating a bi-directional interpretive process.

This social component appears to be a critical factor in determining whether an external symbol or meaning is accepted and integrated into a participant's meaning structure. This finding suggests that human-AI meaning co-construction may be most effective in social contexts where multiple perspectives interact, rather than in isolated human-AI dyads, and bi-directional interpretive processes on the symbols is the key to its success.

\paragraph{Conflict as a Mechanism for Deep Meaning-Making.}
Our analysis reveals that conflict is not a conversational failure but rather an engine/ opportunity for co-construction of meaning. When the CA introduces meanings or symbols that potentially conflict with meanings that participants had  (e.g., ``Your partner wants a loud party, you want a quiet one''), it forces participants to pause, reflect, and articulate their own meanings and symbols in greater detail. This process of meaning-making, especially when meanings are challenged, represents where the deepest meaning-making occurs.


These findings suggest several important considerations for designing human-AI interactions. First, designers should embrace productive conflict by creating CA systems that can introduce meaningful conflicts to stimulate deeper meaning exploration. Second, leveraging social contexts through multi-agent or human-mediated CA interactions can amplify meaning construction effects. Third, supporting the iterative meaning-making process by creating interaction patterns that allow for multiple cycles of interpretation and refinement enhances the depth of meaning co-construction. Finally, recognizing CA as a social actor means designing with the understanding that CA can actively participate in meaning construction rather than merely providing information, fundamentally shifting how we conceptualize human-AI collaborative meaning-making processes.

\subsection{Limitations and Future Work}
\subsubsection{Methodological Limitations}
Our study design presents several constraints that future research should address. First, our participant pool was mostly young adults (18-34) from university settings, potentially missing important variations in the meaning-making process across age groups, educational backgrounds, and cultural contexts. Older adults or individuals with different technological literacy \cite{akello2024digital} might engage with AI-mediated meaning construction quite differently, particularly given varying levels of anthropomorphize and trust in AI systems.

Second, our controlled experimental scenarios, while necessary for observing specific meaning-making dynamics, may not fully capture the complexity of naturalistic human-AI interactions. The pre-scripted nature of Study 1 and the LLM-powered structure of Study 2, though methodologically rigorous, potentially constrained the organic emergence of meaning that might occur in unstructured conversations. Future work should explore longitudinal, in-the-wild studies where meaning co-construction unfolds over extended periods without experimental constraints.

Third, our focus on two specific domains (activity planning and party organization) limits the generalization of our findings. Meaning-making process might differ substantially in high-stakes contexts like medical decision-making, creative collaboration, or educational settings where the consequences of misaligned meanings re more significant.

Fourth, while symbolic interactionism provided a valuable lens for understanding human-AI meaning-making, our application represents just one theoretical perspective. Future research should explore complementary frameworks such as distributed cognition, activity theory, or phenomenological approaches to provide a more comprehensive understanding of human-AI meaning co-construction. Additionally, our work primarily examined dyadic human-AI interactions; future studies should investigate multi-party scenarios where multiple humans and AI agents negotiate meaning, potentially revealing new dynamics of symbolic convergence and divergence.

\subsubsection{Ethical Considerations and Statement.}
Our studies were conducted under IRB approval (Redacted for anonymity) from (Redacted for anonymity). Informed consent was obtained from all participants, who were informed of their right to withdraw at any time. In Study 2, which employed a large language model, automated guardrails were implemented to prevent harmful behavior, and researchers were on site to monitor and resolve any issues. No major ethical issues were reported during the research process. All participant data were anonymized.

Our work raises important ethical questions that warrant deeper investigation. The AI's demonstrated ability to shift human meanings through social framing could be exploited for manipulation or persuasion. Future research must establish ethical guidelines for AI-mediated meaning construction, particularly in vulnerable populations or high-stakes decisions. Additionally, questions of meaning ownership and agency emerge: when meanings are co-constructed with AI, who has authority over the resulting interpretations?

\section{Conclusion}

This paper presents empirical evidence that human-AI interaction can be understood and enhanced through the lens of symbolic interactionism. Our two studies demonstrate that meaning in human-AI conversation is not transferred but actively co-constructed through iterative processes of meaning-making (e.g., meaning introduction, meaning explanation, or symbol consolidation). These findings challenge prevailing assumptions about CAs as passive information processors and reveal their potential as active participants in meaning-making.

Our key contributions include: (1) empirical validation that all three premises of symbolic interactionism operate effectively in human-AI contexts; (2) evidence that productive conflict, rather than immediate agreement, drives deeper meaning construction; (3) demonstration that the AI's influence on meaning is substantially amplified when presented alongside social perspectives; and (4) identification of processes where humans and AI co-construct meanings.

Our findings suggest that future AI systems should be designed not as repositories of fixed definitions but as active partners in the ongoing project of making sense of the world through symbols. By embracing the dynamic, contested, and fundamentally social nature of meaning, we can create AI systems capable of genuine collaboration with humans in our complex and symbol-rich world.

\bibliographystyle{ACM-Reference-Format}
\bibliography{references}

\appendix


\subsection{Studies}

\begin{table}[ht]
\centering
\caption{Participants of Study 1}
\label{tab:participants_study1}
\resizebox{\columnwidth}{!}{%
\begin{tabular}{p{0.12\linewidth} p{0.12\linewidth} p{0.15\linewidth} p{0.25\linewidth} p{0.2\linewidth}}
\toprule
\textbf{Pid} & \textbf{Age} & \textbf{Gender} & \textbf{Race \& Ethnicity} & \textbf{Use of LLM} \\ \midrule
Std1\_P1  & 18–24 & Man   & Other & Twice a week \\
Std1\_P2  & 18–24 & Woman & Asian or Pacific Islander & Everyday \\
Std1\_P3  & 18–24 & Man   & Asian or Pacific Islander & Everyday \\
Std1\_P4  & 18–24 & Man   & Hispanic or Latino & Everyday \\
Std1\_P5  & 18–24 & Man   & Asian or Pacific Islander & Twice a week \\
Std1\_P6  & 18–24 & Woman & Hispanic or Latino & Everyday \\
Std1\_P7  & 18–24 & Woman & Hispanic or Latino & Everyday \\
Std1\_P8  & 18–24 & Woman & Asian or Pacific Islander & Everyday \\
Std1\_P9  & 18–24 & Woman & Asian or Pacific Islander & Once a month \\
Std1\_P10 & 18–24 & Man   & Prefer not to say & Everyday \\
Std1\_P11 & 18–24 & Woman & Asian or Pacific Islander & Twice a week \\
Std1\_P12 & 18–24 & Man   & Asian or Pacific Islander & Everyday \\
Std1\_P13 & 18–24 & Woman & Asian or Pacific Islander & Everyday \\
Std1\_P14 & 18–24 & Woman & Asian or Pacific Islander & Once a week \\
Std1\_P15 & 18–24 & Woman & Other & Never \\
Std1\_P16 & 18–24 & Man   & Asian or Pacific Islander & Twice a week \\
Std1\_P17 & 18–24 & Woman & Hispanic or Latino & Once a week \\
Std1\_P18 & 18–24 & Man   & White or Caucasian & Twice a week \\
Std1\_P19 & 18–24 & Woman & White or Caucasian & Everyday \\
Std1\_P20 & 18–24 & Man   & Asian or Pacific Islander & Once a month \\ 
\bottomrule
\end{tabular}}
\end{table}

\begin{table}[ht]
\centering
\caption{Participants of Study 2}
\label{tab:participants_study2}
\resizebox{\columnwidth}{!}{%
\begin{tabular}{p{0.12\linewidth} p{0.12\linewidth} p{0.15\linewidth} p{0.25\linewidth} p{0.2\linewidth}}
\toprule
\textbf{Pid} & \textbf{Age} & \textbf{Gender} & \textbf{Race \& Ethnicity} & \textbf{Use of LLM} \\ \midrule
Std2\_P1  & 25–34 & Woman & White or Caucasian & Everyday \\
Std2\_P2  & 25–34 & Woman & White or Caucasian & Everyday \\
Std2\_P3  & 25–34 & Man   & Asian or Pacific Islander & Twice a week \\
Std2\_P4  & 18–24 & Woman & Asian or Pacific Islander & Everyday \\
Std2\_P5  & 25–34 & Man   & Prefer not to say & Twice a week \\
Std2\_P6  & 25–34 & Woman & White or Caucasian & A few times \\
Std2\_P7  & 25–34 & Man   & Asian or Pacific Islander & Everyday \\
Std2\_P8  & 35–44 & Woman & Middle Eastern & Twice a week \\
Std2\_P9  & 35–44 & Man   & White or Caucasian & Everyday \\
Std2\_P10 & 35–44 & Man   & Asian or Pacific Islander & Once a month \\
Std2\_P11 & 25–34 & Man   & Asian or Pacific Islander & Everyday \\
Std2\_P12 & 25–34 & Man   & Asian or Pacific Islander & A few times \\
Std2\_P13 & 25–34 & Woman & Asian or Pacific Islander & A few times \\
Std2\_P14 & 25–34 & Woman & Asian or Pacific Islander & Everyday \\
Std2\_P15 & 25–34 & Man   & Asian or Pacific Islander & Once a month \\
Std2\_P16 & 25–34 & Woman & Asian or Pacific Islander & Everyday \\ 
\bottomrule
\end{tabular}}
\end{table}

\begin{table}[h]
\centering
\caption{Post-Survey Questions of Study 1}
\label{tab:post_survey_study1}

\resizebox{\textwidth}{!}{%
\begin{tabular}{p{0.15\linewidth} p{0.75\linewidth} p{0.1\linewidth}}
\hline
\textbf{Part} & \textbf{Question} & \textbf{Response Type} \\
\hline
\multicolumn{3}{l}{\textbf{Part 1: Quick Ratings (1 = Strongly Disagree, 7 = Strongly Agree)}} \\
1.1 & I felt I was actively defining the criteria for what makes a [object/activity] [subjective quality]. & Likert 1–7 \\
1.2 & The AI helped me clarify what [subjective quality] means for me in the context of this specific [object/activity]. & Likert 1–7 \\
1.3 & The interaction felt too superficial to truly explore my personal definition of a [subjective quality] [object/activity]. & Likert 1–7 \\
1.4 & The AI’s suggestions made me reconsider my personal criteria for what makes a [object/activity] [subjective quality]. & Likert 1–7 \\
\hline
\multicolumn{3}{l}{\textbf{Part 2: Your Experience in the Scenario (Good/Memorable/Comfortable)}} \\
2.1 & Your Initial Criteria. When you first thought about the meaning of: \newline 
\hspace{1em} Good: [TEXT] \newline
\hspace{1em} Memorable: [TEXT] \newline
\hspace{1em} Comfortable: [TEXT] & Open-ended \\
2.2 & The External Perspective. The AI introduced an external perspective (e.g., family member’s interests, traffic data). How did this new information make you reconsider your criteria? & Open-ended \\
2.3 & The AI's Method. The AI challenged your initial criteria by offering this perspective. What was your impression of the AI (e.g., helpful, intrusive, intelligent, collaborative)? & Open-ended \\
2.4 & Your Final Criteria (The Meaning Shift). Describe your final criteria for what makes a [subjective quality] [object/activity] now, after the conversation. How is this different from your initial ideas? & Open-ended \\
\hline
\multicolumn{3}{l}{\textbf{Part 3: Final Reflection on Meaning}} \\
3.1 & Defining the Concepts. Thinking about your own life, how would you describe the difference between: \newline
\hspace{1em} The Object (the concrete event, like the [object/activity]) \newline
\hspace{1em} The Subjective Meaning (your personal feeling or judgment about that event)? & Open-ended \\
3.2 & The AI's Focus. Based on the conversation, did you feel the AI was interacting more with the Object (the [object/activity]) or your Subjective Meaning (your personal criteria/feelings)? Please explain. & Open-ended \\
\hline
\end{tabular}%
}
\end{table}

\begin{table}[ht]
\centering
\caption{Post-survey of Study 2}
\label{tab:post_survey_study2}
\resizebox{\linewidth}{!}{%
\begin{tabular}{p{0.97\linewidth}}
\toprule
\textbf{Part 1: Defining Good / Memorable} \\ \midrule
After interacting with the AI, when you think about a good activity, what do you mean by ``Good''? \_\_\_\_\_\_\_\_\_\_\_\_\_\_\_\_\_\_\_\_\_\_ \\[0.5em]
After interacting with the AI, when you think about a memorable party, what do you mean by ``Memorable''? \_\_\_\_\_\_\_\_\_\_\_\_\_\_\_\_\_\_\_\_\_\_ \\ \midrule

\textbf{Part 2: Meaning and AI Support (1 = Strongly Disagree, 7 = Strongly Agree)} \\ \midrule
The AI’s hypothetical social situation (e.g., [Scenario 1: a friend who’s new to the activity / Scenario 2: a family member preferring a small, quiet gathering]) helped me rethink what makes [a good activity / a memorable party]. \\[0.5em]
The AI’s perspective on what makes [a good activity / a memorable party] (e.g., [Scenario 1: balancing thrill with a friend’s comfort / Scenario 2: blending energy with intimacy]) felt insightful. \\[0.5em]
The AI’s suggestions in the conversation made me reconsider my personal definition of [good / memorable]. \\[0.5em]
The AI’s final summary (e.g., describing [good / memorable] as [connection / joy]) captured the essence of our conversation. \\ \midrule

\textbf{Part 3: Interaction Dynamics (1 = Strongly Disagree, 7 = Strongly Agree)} \\ \midrule
I felt actively involved in shaping what [good / memorable] means during the conversation. \\[0.5em]
The AI helped me clarify my personal sense of what makes [an activity good / a party memorable]. \\[0.5em]
The conversation felt too shallow to fully explore my idea of [good / memorable]. \\[0.5em]
Did you perceive that the AI followed the meaning of ``Good'' and ``Memorable'' that you mentioned? \\[0.5em]
Did the AI suggest ``a good activity'' and ``a memorable party'' based on the meaning you expressed at the end of the conversation? \\ \midrule

\textbf{Open-Ended Questions} \\ \midrule
Did you feel the AI focused more on the event itself (e.g., [activity details / party plans]) or on your definition of what makes it [good / memorable]? Please explain. \\[0.5em]
Why do you think you perceived that the AI followed your meaning of ``Good'' and ``Memorable''? \\[0.5em]
How was the suggestion that the AI provided at the end of the interaction? \\ \bottomrule
\end{tabular}}
\end{table}

\begin{table}[ht]
\centering
\caption{Conversation counts per PID with AI and Participant turn distribution of Study 2.}
\label{tab:convo_counts}
\setlength{\tabcolsep}{3pt} 
\renewcommand{\arraystretch}{0.9} 
\resizebox{\columnwidth}{!}{%
\begin{tabular}{lccc}
\toprule
\textbf{PID} & \textbf{Total Conversations} & \textbf{AI Turns} & \textbf{Participant Turns} \\
\midrule
Std2\_P1  & 24 & 12 & 11 \\
Std2\_P2  & 17 &  9 &  8 \\
Std2\_P3  & 21 & 10 & 10 \\
Std2\_P4  & 26 & 15 & 11 \\
Std2\_P5  & 23 & 11 & 12 \\
Std2\_P6  & 30 & 14 & 16 \\
Std2\_P7  & 27 & 14 & 12 \\
Std2\_P8  & 22 & 13 &  9 \\
Std2\_P9  & 64 & 39 & 25 \\
Std2\_P10 & 33 & 17 & 14 \\
Std2\_P11 & 34 & 24 &  9 \\
Std2\_P12 & 47 & 23 & 20 \\
Std2\_P13 & 21 & 10 &  8 \\
Std2\_P14 & 14 &  8 &  4 \\
Std2\_P15 & 34 & 13 & 18 \\
Std2\_P16 & 16 &  8 &  4 \\
\bottomrule
\end{tabular}%
}
\end{table}

\end{document}